# Construction of Rough graph to handle uncertain pattern from an Information System

R. Aruna Devi*, K. Anitha**

*Research scholar, Department of Mathematics, SRMIST, Ramapuram, Chennai 600089, India
**Department of Mathematics, SRMIST, Ramapuram, Chennai 600089, India
ORCID: 0000-0002-1316-9185

## Abstract

Rough membership function defines the measurement of relationship between conditional and decision attribute from an Information system. In this paper we propose a new method to construct rough graph through rough membership function $\omega_G^F(f)$. Rough graph identifies the pattern between the objects with imprecise and uncertain information. We explore the operations and properties of rough graph in various stages of its structure.

**Keywords**: *Set approximations; Rough membership function; Rough graph*

**AMS Subject Classification:** *05C72 · 94D05 · 05C90*

## 1 Introduction

The generic and simple way of making optimal decisions is graphical representation of the given problem. Attribute value system is one of the finest methods to make decision analysis. For handling the information system with uncertainty rough set approach is an efficient technique. This theory was constructed by Pawlak through the concept of indiscernibility relation. This involves information system which contains universe set and the attributes both conditional and decision. The objects which are having same attribute values are indiscernible, otherwise they are discernible with each other [1]. The indiscernibility relation is represented in the form of matrix and graph [7]. The matrix representation of discernible relation was introduced by Skowron [1]. By using two crisp sets, the approximation space has been considered. Let X be the subset of universe and if the equivalence class contains in X, then it is said to be lower approximation of X and if the intersection of the equivalence class and X is non empty, then it is said to be an upper approximation of X [26]. There are many properties defined on lower and upper approximations [13].
Instead of rough approximation, rough membership function was proposed [8] to define rough set. Through the definition of rough membership function, the set approximations have been defined. Pawlak has defined two types of information system in rough set theory with and without decision attributes respectively [9]. The minimal representation of the information system is named as reduct [1]. The wide range of applications of the rough set are being implemented in the field of decision making approaches, artificial intelligence, Machine and deep larning techniques.
The rough graph is a mechanism of embedding graph in rough set domain. In 2006, He Tong and Kai Shi gave the structure for rough graph and extended as weighted rough graph by giving class weights for edges [22]. He Tong gave COTA algorithm for finding the shortest path through Kruskal and Dijkstra's algorithm for rough optimal tree. The edge rough graph was constructed by the partition of edge set [23]. Bibin Mathew defined vertex rough graph by dividing the vertex set [6]. Shu defined a Rough graph model with a double universe of discourse [11].
Rough Graph was represented interms of matrix and edge list [19]. The properties of rough graph interms of similarity degree, precision etc. are defined in [6,20,21] . In this paper, we have introduced the novel idea for constructing rough graph through rough membership function. Section 2, deals about the basic concepts of rough sets with existing graph structure. In section 3, we brief out the construction process of rough graph followed by the properties of rough walk, rough path and rough cycle. In section 5, we discuss about the operations on rough graph and its properties. At last, we exhibit concluded remarks with future work.

## *2* **Preliminaries**

**Definition *1*.[1]:** Let U be the universe set and an equivalence relation $B \subseteq U \times U$. Let Y be a subset of U. The lower and upper approximations of Y are given as

$$B_\nabla(Y) = \bigcup_{x \in Y} \{B(x): B(x) \subseteq Y\}$$

$$B^\nabla(Y) = \bigcup_{x \in Y} \{B(x): B(x) \cap Y \neq \phi\}$$

**B**- boundary region of Y

$$BN_B(Y) = B^\Delta(Y) - B_\nabla(Y)$$

*The pair $(B_\nabla(Y), B^\Delta(Y))$ is called a Rough set.*

**Definition 2.**[22] : Let $U = (V, E)$ be the universe graph with the universe of discourse $U = \{e_1, e_2, \ldots, e_{|U|}\}$ having the attribute set on U is $R = \{r_1, r_2, \ldots, r_{|R|}\}$, where $V = \{v_1, v_2, \ldots, v_n\}$, $E = \cup e_k(v_i, v_j)$. The elements in E divided into different equivalence class $[e]_R$ from the attribute set $R \subseteq \mathbb{R}$ on E. By two exact graphs $\underline{R}(T) = (W, (X))$ which is lower approximation of X and $\overline{R}(T) = (W, \overline{R}(X))$ which is upper approximation of X can be used to define it approximately, where

$$\underline{R}(X) = \{e \in E \mid [e]_R \subseteq X\}$$

$$\overline{R}(X) = \{e \in E \mid [e]_R \cap X \neq \phi\}$$

The pair $(\overline{R}(T), \overline{R}(T))$ is called R−rough graph.

**Definition 3** [22]: Let $U = (V, E)$ be universe graph, where $V = \{v_1, v_2, \ldots, v_n\}$, $E = \cup e_k(v_i, v_j)$, for every $e \in E$ with the mapping for the edge weight is $\omega: e \rightarrow \omega(e)$. The class weights for the edge equivalence class is given as $\omega[e_{uv}]_R = f(\omega(e))$, where $[e_{uv}]_R$ − edge equivalence class between the vertex u and vertex v respect to attribute R. The class weight of their edge equivalence class for the rough graph $T = (\underline{R}(T), \overline{R}(T))$ is called weighted rough graph.

**Definition 4**. [23]: Let $\mathcal{A} = (K, R)^e$ be an edge approximation space. Given an edge subset $X \subseteq E(K)$, be the lower and upper approximation of X in $\mathcal{A}$, denoted by $\underline{Q}(X)$ and $\overline{Q}(X)$ respectively and defined as

$$\underline{Q}(X) = \{x \in E(K) \mid J_Q(x) \subseteq X\}$$

$$\overline{Q}(X) = \{x \in E(K) \mid J_Q(x) \cap X \neq \phi\}$$

**Definition 5**. [6]: By two exact graphs, R −vertex rough graph is defined as $R_\nabla(H) = (R_\nabla(K), R_\nabla(L))$ and $R^\Delta(H) = (R^\Delta(K), R^\Delta(L))$, where

$$R_\nabla(K) = \{v \in V : [v]_R \subseteq K\}$$

$$R^\Delta(K) = \{v \in V : [v]_R \cap K \neq \phi\}$$

$$R_{\bar{V}}(L) = \{(v_i, v_j) \in L : v_i, v_j \in [v]_R \text{ for some } v \in R_{\bar{V}}(K)\}$$

$$R^{\Delta}(L) = \begin{cases} (v_i, v_j) \in E & : v_i \in [v_i]_R \text{ and } [v_i]_R \cap L \neq \phi\} \\ v_i & : [v_i]_R \text{ and } [v_j]_R \cap L \neq \phi. \end{cases}$$

**Definition 6.**[8]:Assume $\mathbb{M} = (U, F)$ is an information system and $\phi \neq G \subseteq U$. The Rough membership function for the set G is

$$\omega_G^F(f) = \frac{|[f]_G \cap G|}{|[f]_F|} \text{ for some } f \in U$$

**Example 1**. Let $\mathbb{M} = (U, F)$ be an information system. Let $U = \{1,2,3,4,5,6,7,8,9,10,11,12,13,14,15\}$ and $G = \{1,2,3,4,5,6,7,8\}$. Let thirst, hunger, frequent, weight loss, tiredness are conditional attributes and diabetic is the decision attribute. The decision system is given as follows.

**Table 1**:Decision system

| Patients | Thirst | Hunger | Frequent | Weight Loss | Tiredness | Diabetic |
|---|---|---|---|---|---|---|
| $P_1$ | H | H | L | L | H | H |
| $P_2$ | H | H | L | L | L | H |
| $P_3$ | H | H | H | L | H | H |
| $P_4$ | H | H | H | L | L | H |
| $P_5$ | H | L | H | H | H | H |
| $P_6$ | H | H | H | H | H | H |
| $P_7$ | H | L | L | L | L | H |
| $P_8$ | H | H | H | H | H | H |
| $P_9$ | H | H | L | L | H | L |
| $P_{10}$ | H | L | H | L | H | L |
| $P_{11}$ | H | H | H | L | H | L |
| $P_{12}$ | H | L | L | L | L | L |
| $P_{13}$ | L | H | L | H | H | L |
| $P_{14}$ | L | L | L | H | L | L |
| $P_{15}$ | H | H | L | L | H | L |

The rough membership values are,

$$\omega_G^F(1) = \frac{|\{1,9,15\} \cap \{1,2,3,4,\ldots,8\}|}{|\{1,9,15\}|} = \frac{1}{3}; \; \omega_G^F(2) = 1$$

Similarly we can find other membership values.

# 3.Rough Graph and its Properties

*In this section, we have constructed the rough graph from the information system through the rough membership function.*

**Definition 7**. *Let $\mathfrak{R} = \{V, E, \omega\}$ be a triple consisting of non-empty set $V = \{v_1, v_2, \ldots, v_n\} = U$, where U is a universe, $E = \{e_1, e_2, \ldots, e_m\}$ be an edge set for V and $\omega$ be a function $\omega: V \to [0,1]$. A rough graph is defined as,*

$$\Re(v_i, v_j) = \begin{cases} \text{If } \max\left(\omega_G^V(v_i), \omega_G^V(v_j)\right) > 0, & \text{then } v_i v_j \text{ exists} \\ \text{If } \max\left(\omega_G^V(v_i), \omega_G^V(v_j)\right) = 0, & \text{no edge.} \end{cases}$$

**Remark 1.** Rough graph is always simple and undirected graph.

**Example 2.** From example Example 1; the rough graph is constructed as shown below

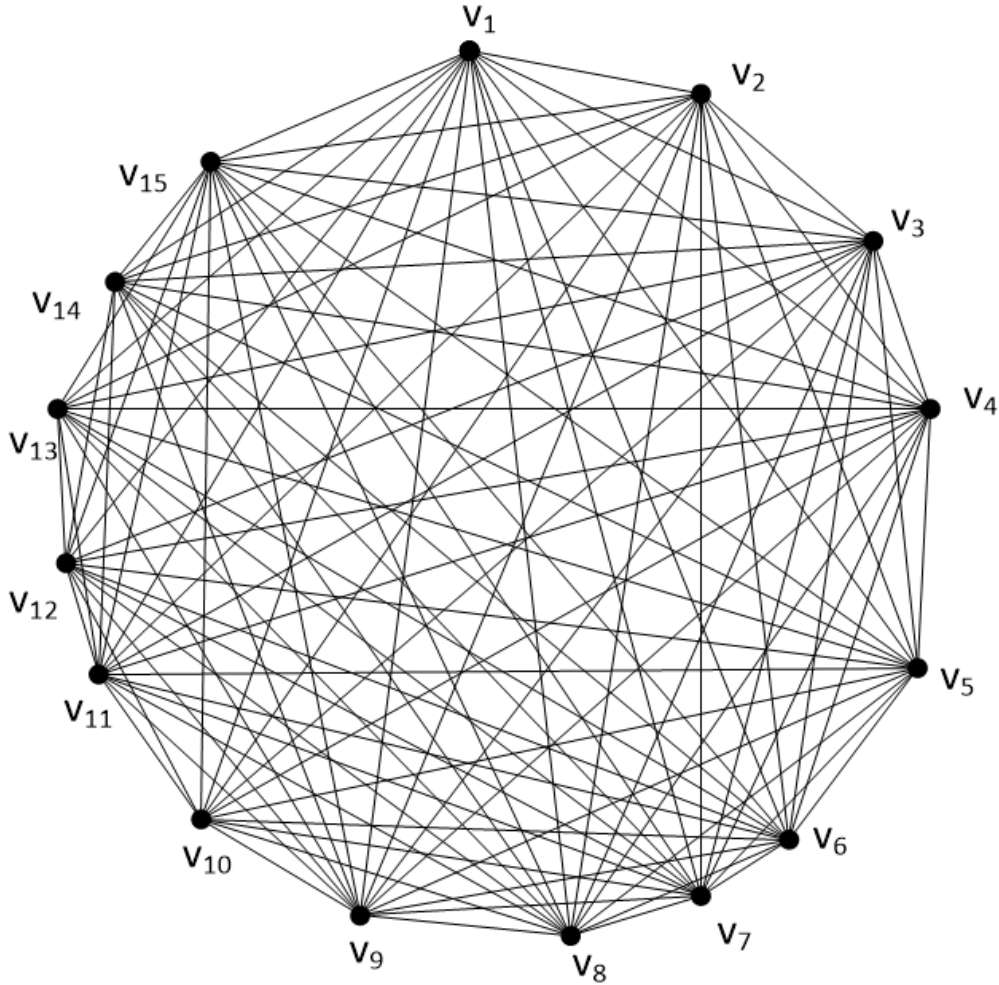

**Figure 1**:Rough graph

**Definition 8.** The degree of a vertex $v_i$ of rough graph $\Re$ is defined as the number of edges incident to that vertex. It is denoted by $\Delta_\Re$.

**Example 3.** From example 2, the degree of vertex in rough graph will be

$$\Delta_\Re(v_1) = 14$$

Similarly we can find the degree for other vertices.

**Theorem 1.** Rough graph is always a connected and pendant free graph.

Proof. In rough graph $\forall v_i, \Delta_\Re(v_i) > 1$, proof follows.

**Remark 2.**

1. Rough graph satisfies hand-shaking lemma.
2. In any rough graph, the number of vertices of odd degree is always even.

**Definition 9.**

$$+(v) = \max\{\Delta_\Re(v_i)|v_i \in V(G)\}$$

$$-(v) = \min\{\Delta_\Re(v_i)|v_i \in V(G)\}$$

If $+(v) = -(v)$, then the graph is said to be regular rough graph.

**Definition 10.** If every vertex of rough graph is adjacent with all other vertices, then it is said to be complete rough graph. The complete rough graph with n vertices is denoted by $\Im_n$.

**Example 4.** By using the following information system, the complete rough graph can be constructed.

**Table 2**:Information system

| Subjects | Assignment | Test | Internal | Result |
|---|---|---|---|---|
| $x_1$ | Not Submitted | Written | >80 | Pass |
| $x_2$ | Submitted | Not Written | >80 | Pass |
| $x_3$ | Submitted | Written | >90 | Pass |
| $x_4$ | Not Submitted | Written | >50 | Fail |
| $x_5$ | Submitted | Not Written | >50 | Fail |
| $x_6$ | Not Submitted | Written | >90 | Fail |

The complete rough graph for the table 2 is represented in figure 2

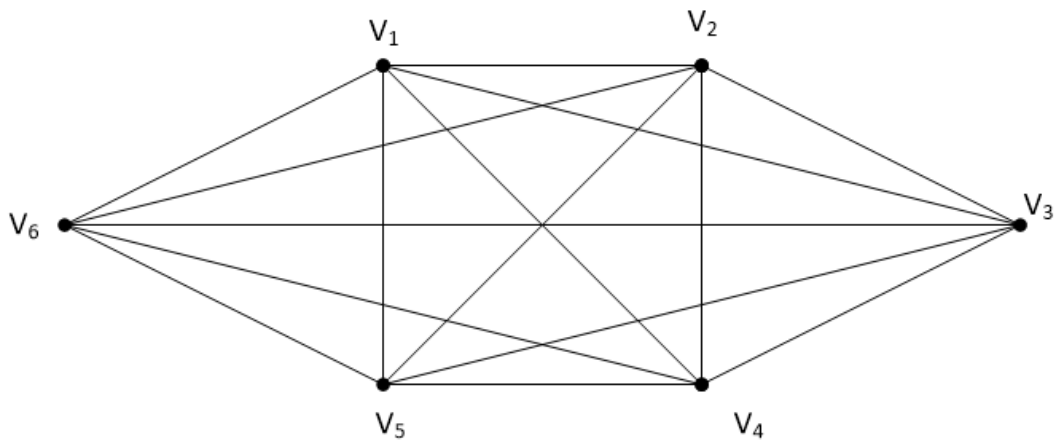

**Figure 2**

**Remark 3.**

1. A complete rough graph with n vertices contains $\frac{n(n-1)}{2}$ edges.
2. A complete rough graph is always $(n-1)$ regular rough graph.

# 4. Rough walk and Rough cycle

**Definition 11.** An alternating sequence of vertices and edges in a rough graph with $\omega(v_i) \geq 0.5$, is said to be rough walk.

**Example 5.** The decision system is given as,

Table 3: Decision system

| Section | Inside | Outside | Difference | Churn |
|---|---|---|---|---|
| 1 | Middle | Middle | Small | False |
| 2 | Large | Large | Small | False |
| 3 | Small | Small | Small | False |
| 4 | Small | Small | Large | True |
| 5 | Middle | Middle | Small | True |
| 6 | Middle | Small | Small | True |

The rough walk for the above decision table is $v_1, v_2, v_4, v_5, v_2, v_6, v_3$

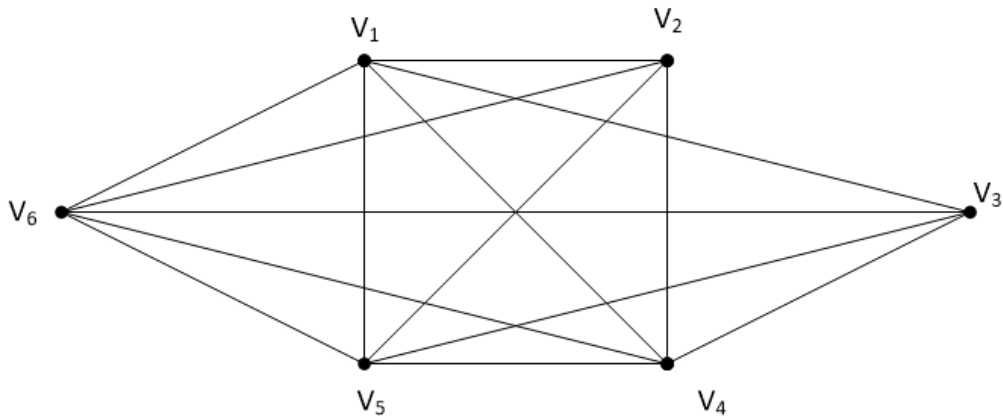

**Figure 3**

**Definition 12.** Distinct edges in rough walk, is said to be rough trail and distinct vertices in a rough walk, is said to be rough path. It is denoted by $\mathfrak{P}_n$.

**Example 6.** From Example 5; the rough trail and rough path is given as, $v_1, v_2, v_4, v_5, v_3, v_5$ is a rough trail.

$$v_1, v_3, v_6, v_4, v_5, v_2 \text{ is a rough path.}$$

**Definition 13.** A rough cycle is defined as the closed rough walk $v_1, v_2, \ldots, v_n = v_1$, where $n \geq 3$ and $v_1, v_2, \ldots, v_{(n-1)}$ are distinct. It is denoted by $\mathfrak{C}_n$.

**Example 7.** From Example 5 rough cycle is constructed as follows,

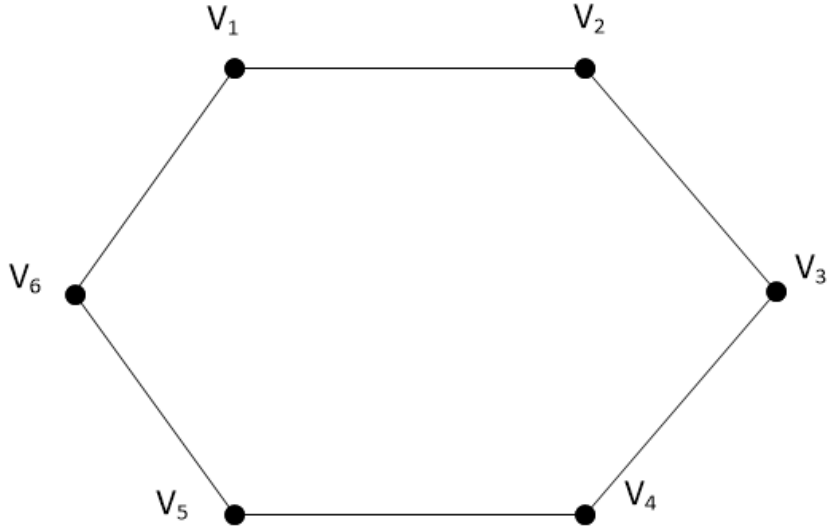

**Figure 4**

**Theorem 2.** In a rough graph, any $v_1 - v_2$ rough walk contains at-least one $v_1 - v_2$ rough path.

Proof. Since rough walk contains repeated vertices distinct rough paths can be developed in $v_1 - v_2$
∴ Any $v_1 - v_2$ rough walk contains $v_1 - v_2$ rough path.

## 5 Operations on Rough graph

**Definition 14.** Let $\Re_1(V_1, E_1)$ and $\Re_2(V_2, E_2)$ be two rough graphs with $V_1 \cap V_2 = \phi$. Then rough union of $\Re_1$ and $\Re_2$ is defined as $\Re_1 \cup \Re_2 = (V_1 \cup V_2, E_1 \cup E_2)$, where

$$V_1 \cup V_2(x) = \begin{cases} \omega_1(x), & \text{if } x \in V_1 \\ \omega_2(x), & \text{if } x \in V_2 \end{cases}$$

**Definition 15.** Let $\Re_1(V_1, E_1)$ and $\Re_2(V_2, E_2)$ be two rough graphs with $V_1 \cap V_2 = \phi$. Then the rough join of $\Re_1$ and $\Re_2$ is defined as joining all vertices of $V_1$ to the vertices of $V_2$.
(i.e) $\Re_1 \sim \Re_2 = (V_1 \sim V_2, E_1 \sim E_2)$ where,

$$V_1 \sim V_2(x) = \begin{cases} \omega_1(x), & \text{if } x \in V_1 \\ \omega_2(x), & \text{if } x \in V_2 \end{cases}$$

**Example 8.** Let $\Re_1$ and $\Re_2$ be two rough graphs,

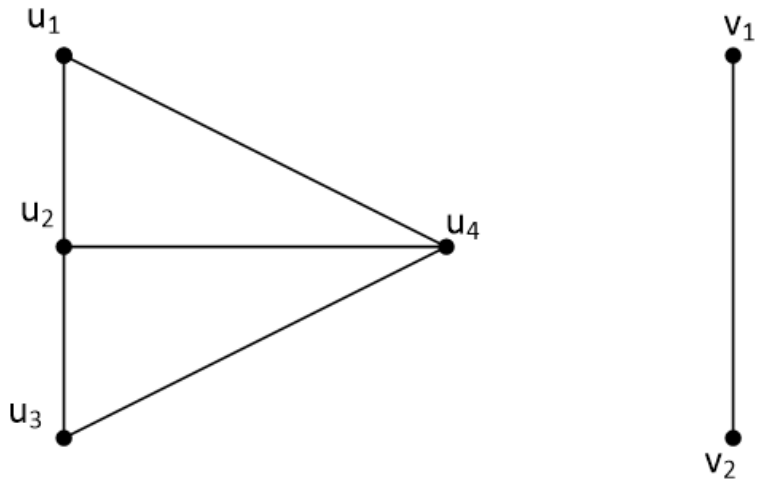

**Figure 5:** $\Re_1$ *and* $\Re_2$

The rough join of $\Re_1$ and $\Re_2$ is,

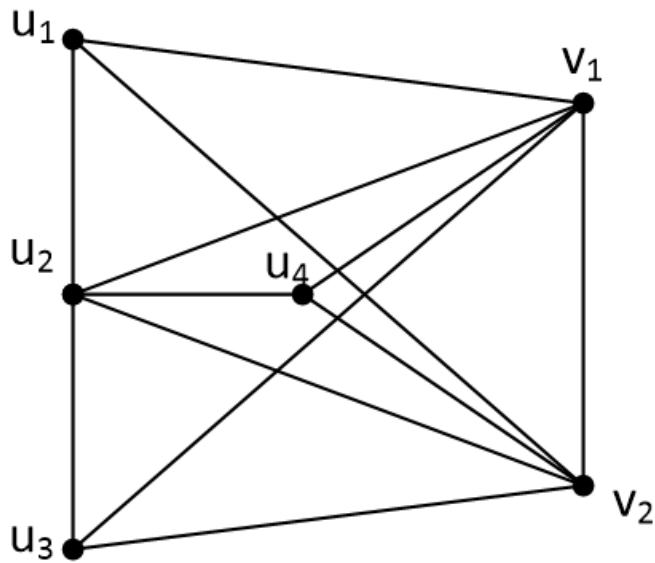

**Figure 6:** $\Re_1 \sim \Re_2$

**Definition 16.** Let $\Re_1 = (V_1, E_1)$ and $\Re_2 = (V_2, E_2)$ be two rough graphs with the vertex sets as $V_1$ and $V_2$ and their edge sets as $E_1$ and $E_2$ respectively. Then the rough cartesian product of two rough graphs $\Re_1$ and $\Re_2$ are defined as

$$V_1 \approx V_2 = \{(u,v) | u \in V_1 \text{ and } v \in V_2\}$$

$$E_1 \approx E_2 = \{(u,v)(x,y) | \omega(u) = \omega(x), vy \in E_2 \text{ or } ux \in E_1, \omega(v) = \omega(y)\}$$

with $(\omega_1 \approx \omega_2)(u,v) = \omega_1(u) \wedge \omega_2(v)$

**Example 9.** Let $\Re_1$ and $\Re_2$ be two rough graphs are defined below

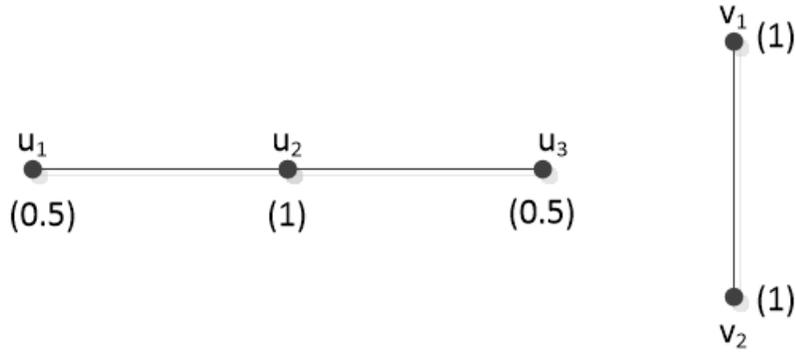

**Figure 7:** $\Re_1$ and $\Re_2$

$V(H) \approx V(G) = \{(u_1, v_1), (u_1, v_2), (u_2, v_2), (u_2, v_1), (u_3, v_1), (u_3, v_2)\}$
The rough cartesian product of $\Re_1 \approx \Re_2$ will be

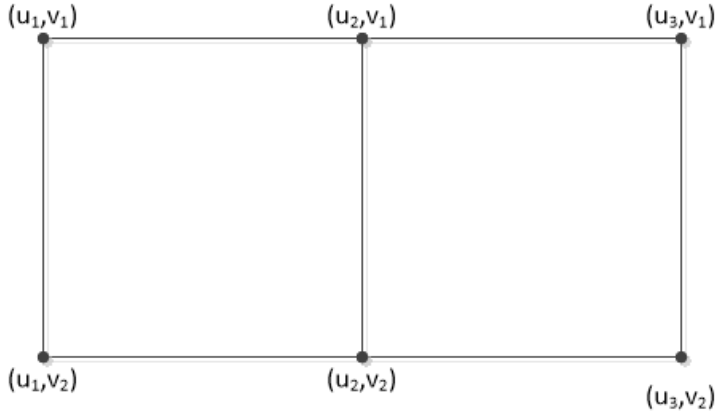

**Figure 8:** $\Re_1 \approx \Re_2$

**Definition 17.** The degree of a vertex in a rough cartesian product for rough graphs is defined as the sum of all edges to a vertex $(u_i, v_i)$ such that

$$\Delta_{\Re_1 \approx \Re_2} = \omega(u_i) \wedge \omega(v_i) \wedge \omega(v_j), \text{ if } u_i = u_j, v_i = v_j$$

**Example 10.** By Example 9, we can find the degree of a vertex for the rough cartesian product is

$$\Delta(u_1, v_1) = 0.5 + 0.5 = 1$$

Simillary we can find for the other vertices.

**Definition 18.** The grid rough graph is defined as the rough cartesian product of two rough paths. It is denoted by $\mathfrak{G}_{m \times n}$.

**Theorem 3.** The rough cartesian product of two rough path is a grid rough graph.

Proof. The rough cartesian product on two rough graphs is defined as

$$V_1 \approx V_2 = \{(u, v) | u \in V_1 \text{ and } v \in V_2\}$$

$$E_1 \approx E_2 = \{(u,v)(x,y) \mid \omega(u) = \omega(x), vy \in E_2 \text{ or } ux \in E_1, \omega(v) = \omega(y)\}$$

Since by the rough path, the membership values between two vertices will be $\geq 0.5$.

Case 1: $m = n$
Then

$$V(\mathfrak{R}_1) \approx V(\mathfrak{R}_2) = \{(u_1, v_1), (u_1, v_2), \ldots, (u_1, v_n), (u_2, v_1), (u_2, v_2), \ldots, (u_2, v_n), \ldots, (u_n, v_1), (u_n, v_2), \ldots, (u_n, v_n)\}$$

$$E(\mathfrak{R}_1) \approx E(\mathfrak{R}_2) = \{\{(u_j, v_i), (u_j, v_{i+1})\} \cup \{(u_i, v_j)(u_{i+1}, v_j)\}\} \quad , 1 \leq i \leq n-1, j = 1,2,3 \ldots$$

By using $V(\mathfrak{R}_1) \approx V(\mathfrak{R}_2)$ and $E(\mathfrak{R}_1) \approx E(\mathfrak{R}_2)$, the resulting graph will be n × n-grid rough graph.

Case 2: $m \neq n$
The number of vertices for the rough cartesian product will be,

$$V(\mathfrak{R}_1) \approx V(\mathfrak{R}_2) = \{(u_1, v_1), \ldots, (u_1, v_n), \ldots, (u_n, v_1), \ldots, (u_n, v_n)\}$$

$$E(\mathfrak{R}_1) \approx E(\mathfrak{R}_2) = \{\{(u_m v_i)(u_m v_{i+1})\} \cup \{(u_i v_n)(u_{i+1} v_n)\}\}$$

By using $V(\mathfrak{R}_1) \approx V(\mathfrak{R}_2)$ and $E(\mathfrak{R}_1) \approx E(\mathfrak{R}_2)$, the resulting graph will be m × n-grid rough graph.

**Example 11.**

$$V(\mathfrak{P}_4) \approx V(\mathfrak{P}_3) = \begin{cases} (u_1, v_1), (u_1, v_2), (u_1, v_3), (u_2, v_1), (u_2, v_2), (u_2, v_3), (u_3, v_1), (u_3, v_2), \\ (u_3, v_3), (u_4, v_1), (u_4, v_2), (u_4, v_3), (u_4, v_4) \end{cases}$$

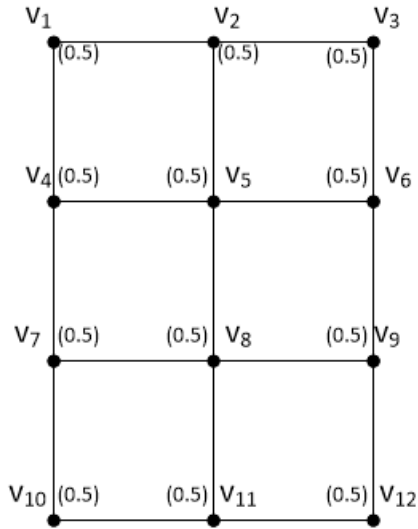

**Figure 9:** $\mathfrak{G}_{4,3}$

**Definition 19.** The ladder rough graph is defined as the rough cartesian product of rough path and the complete rough graph $\mathfrak{I}_2$. It is denoted by $\mathfrak{L}_m$.

**Theorem 4.** For the rough cartesian product of $\mathfrak{I}_2$ and $\mathfrak{P}_n$, $n \geq 2$, the resulting graph will be rough ladder graph.

Proof. By the definition of rough cartesian product, the vertex set of $\mathfrak{I}_2$ and $\mathfrak{P}_n$ will be $V(\mathfrak{I}_2) \approx V(\mathfrak{P}_n) = \{(u_1, v_1), \ldots (u_1, v_n), (u_2, v_1), \ldots, (u_2, v_n)\}$

The number of elements in the vertex set $V(G) \approx V(H)$ will be $2 \times n$. The edges between $\mathfrak{I}_2 \approx \mathfrak{P}_n$ will be

$E(\mathfrak{I}_2) \approx E(\mathfrak{P}_n) = \{(u_1, v_j)(u_2, v_j) \cup (u_1, v_i)(u_1, v_{i+1}) \cup (u_2, v_i)(u_2. v_{i+1})\}\, j = 1,2;\, i = 1,2.\, n$

Then the ladder rough graph $(L_{(n-1)})$ is obtained from the vertex set and the edge set.

**Example 12.** The $\mathfrak{I}_2$ and $\mathfrak{P}_5$ is given as [5.10]

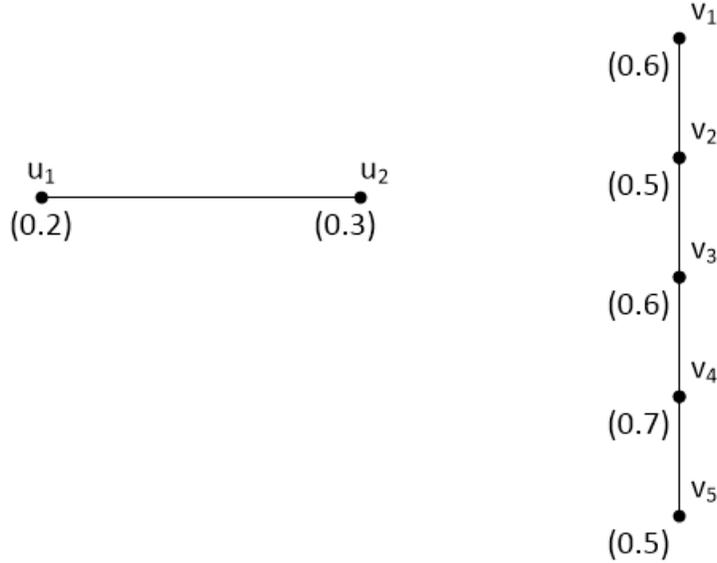

**Figure 10:** $\mathfrak{I}_2$ *and* $\mathfrak{P}_5$

$V(\mathfrak{I}_2) \approx V(\mathfrak{P}_5)$
$$= \begin{cases}((u_1, v_1), (u_1, v_2), (u_1, v_3), (u_2, v_1), (u_2, v_1), (u_2, v_2), (u_2, v_3), (u_3, v_1), (u_3, v_2),\\ (u_3, v_3), (u_4, v_1), (u_4, v_2), (u_4, v_3))\end{cases}$$

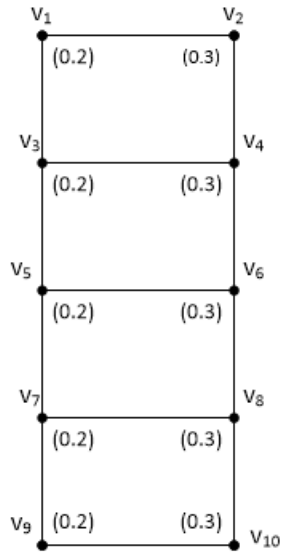

**Figure 11:** $\mathcal{L}_4$

# 6 Conclusion

In this paper we have proposed the constructional technique of Rough graph with the help of rough membership function. Also extensive properties of rough graph are briefly discussed in this paper. In our future work we planned to propose the properties of rough metric dimension for the purpose of attribute reduction.